\documentclass{ecai}
\usepackage[T1]{fontenc}
\usepackage[utf8]{inputenc}

\usepackage{times}
\usepackage{graphicx}
\usepackage{latexsym}
\usepackage{hyperref}
\usepackage{booktabs}
\usepackage{multirow}
\usepackage{tabularx}
\usepackage{float}
\usepackage{color,soulutf8}
\usepackage{adjustbox}

\newcolumntype{Y}{>{\centering\arraybackslash}X}

\begin{document}

\title{El Departamento de Nosotros: How Machine Translated Corpora Affects Language Models in MRC Tasks}

\author{Maria Khvalchik\institute{SemanticWebCompany, Austria,  maria.khvalchik@semantic-web.com} \and Mikhail Galkin \institute{TU Dresden \& Fraunhofer IAIS,
Germany, mikhail.galkin@tu-dresden.de} }

\maketitle
\bibliographystyle{ecai}

\begin{abstract}
  Pre-training large-scale language models (LMs) requires huge amounts of text corpora. 
  LMs for English enjoy ever growing corpora of diverse language resources. 
  However, less resourced languages and their mono- and multilingual LMs often struggle to obtain bigger datasets.
  A typical approach in this case implies using machine translation of English corpora to a target language.
  In this work, we study the caveats of applying directly translated corpora for fine-tuning LMs for downstream natural language processing tasks and demonstrate that careful curation along with post-processing lead to improved  performance and overall LMs robustness.
  In the empirical evaluation, we perform a comparison of directly translated against curated Spanish SQuAD datasets on both user and system levels.
  Further experimental results on XQuAD and MLQA transfer-learning evaluation question answering tasks show that presumably multilingual LMs exhibit more resilience to machine translation artifacts in terms of the exact match score.
\end{abstract}

\section{INTRODUCTION}
Numerous research studies demonstrate how important the data quality is to the outcomes of neural networks and how severely they are affected by low quality data~\cite{DBLP:conf/iq/SessionsV06}.

However, recently transfer learning, where a model is first pre-trained on a data-rich task before being fine-tuned on a downstream task, has emerged as a powerful technique in natural language processing. Models like T5~\cite{DBLP:journals/corr/abs-1910-10683} are now showing human-level performance on most of the well-established benchmarks available for natural language understanding.

Yet, language understanding is not solved, even in well-studied languages like English, even when tremendous resources are used. This paper focuses on a less resourced language, Spanish, and pursues two goals through the lens of open-domain question answering. 

First, we seek to demonstrate that data quality is an important component for training neural networks and overall increases natural language understanding capabilities. Hence, the quality of data should be considered carefully. We consider two recent Neural Machine Translated (NMT) SQuAD datasets, discussed in more detail in Section~\ref{sec:user}, for machine reading comprehension (MRC) in Spanish of different quality, i.e., with and without additional human supevision.

Second, after providing evidence of the data quality difference, we fine-tune both datasets on pre-trained multilingual and monolingual Spanish BERT models and see that there is a significant performance gap in terms of Exact Match (EM) and F1 scores in dev sets of the SQuAD family. However, unexpectedly, the results become more comparable when testing on external benchmarks recently proposed for cross-lingual Extractive QA.

Hence, we tackle the effects of both data quality and neural networks characteristics in an effort to demonstrate that both mentioned above are major factors in the outcomes and should be given equal respect in their primary design. 

\section{RELATED WORK}

Historically, most of NLP tasks, datasets, and benchmarks were created in English, e.g., the Penn Treebank~\cite{DBLP:journals/coling/MarcusSM94}, SQuAD~\cite{DBLP:conf/emnlp/RajpurkarZLL16}, GLUE~\cite{DBLP:conf/iclr/WangSMHLB19}.
Therefore, most of the large-scale pre-trained models were trained in the English-only mode, e.g., BERT~\cite{DBLP:conf/naacl/DevlinCLT19} employed Wikipedia and the BookCorpus~\cite{DBLP:conf/iccv/ZhuKZSUTF15} as training datasets.
Later on, the NLP community sought after increasing language diversity and multilingual models started to appear, such as mBERT or XLM~\cite{DBLP:conf/nips/ConneauL19}.

However, large and diverse enough pre-training corpora of high quality often do not exist.
Several methods have been developed to bridge this gap, e.g., applying machine translation frameworks to English corpora, or performing cross-lingual transfer learning~\cite{DBLP:conf/emnlp/WuD19}. 
Multilingual language models and pre-trained non-English language models are definitely in the focus of the NLP community. 
Still, the language understanding capabilities (hence, the performance) of language models largely depend on data collection and cleaning steps.
In the MRC dimension, for instance, Italian SQuAD~\cite{10.1007/978-3-030-03840-3_29} is obtained via direct translation from the English version whereas French FQuAD~\cite{DBLP:journals/corr/abs-2002-06071} and Russian SberQuAD~\cite{DBLP:journals/corr/abs-1912-09723} have been created based on their language-specific part of Wikipedia often being much smaller than original SQuAD.

With the surge of language-specific pre-trained LMs several benchmarks have been developed that aim at evaluating multi- and cross-lingual characteristics of such LMs. Specifically, for the machine reading comprehension and question answering (QA) task there exist  XQuAD~\cite{DBLP:journals/corr/abs-1910-11856} and MLQA~\cite{DBLP:journals/corr/abs-1910-07475}. 
In this work, we study how LMs perform in QA tasks in Spanish when fine-tuning on datasets of possibly different quality, i.e., directly machine translated and curated with the human-in-the-loop strategy.

\section{PROBLEM FORMULATION}

\begin{figure*}[ht]
    \centering
    \includegraphics[width=\textwidth]{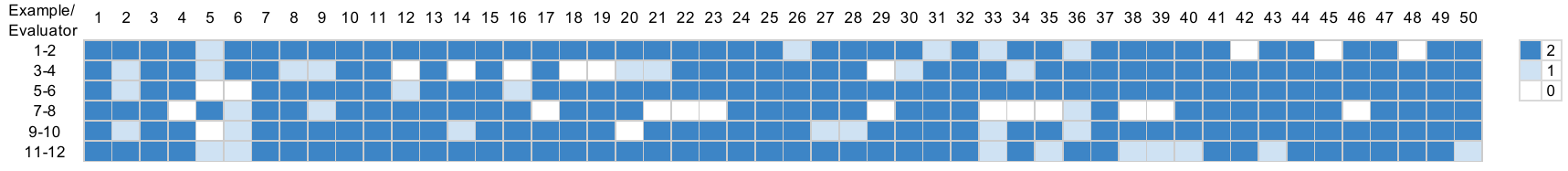}
    \caption{TAR translation evaluation heat map on 50 parallel SQuAD examples scored by 12 evaluators. }
    \label{fig:tar_w}
\end{figure*}

\begin{figure*}[ht]
    \centering
    \includegraphics[width=\textwidth]{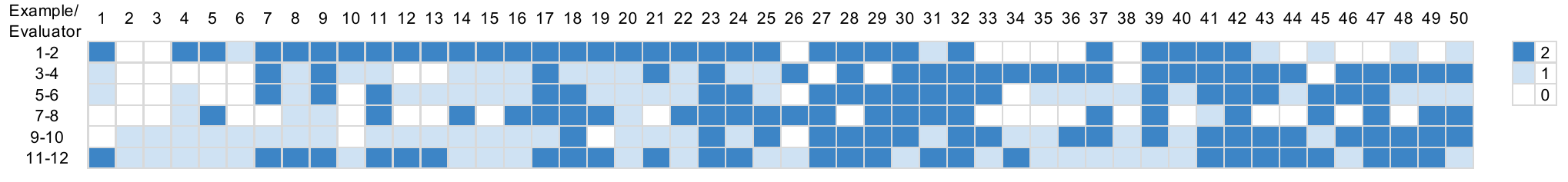}
    \caption{MT translation evaluation heat map on 50 parallel SQuAD examples scored by 12 evaluators.}
    \label{fig:mt_h}
\end{figure*} 

In this work, we aim at exploring the impact of machine translated corpora quality on downstream MRC tasks which is of high importance in less resourced languages. 
We consider Spanish as the target language as one of the most spoken languages in the world that nevertheless has a relatively little amount of available corpora for pre-training modern LMs for language understanding tasks. 
Taking this into account, we tackle the following research questions:
\begin{itemize}
    \item \textbf{RQ1:} Is there a quantifiable difference in data quality between machine translated and manually post-processed corpora for Spanish SQuAD datasets? In order to answer this question, we conduct a user study in Section~\ref{sec:user}.
    \item \textbf{RQ2:} Can we expect the performance difference of LMs in MRC QA tasks when fine-tuned on datasets of different quality? We perform an experimental study in Section~\ref{sec:squad}.
    \item \textbf{RQ3:} Is there a performance difference in downstream transfer-learning QA tasks, i.e., on external benchmarks the LMs were not fine-tuned on? Experimental results are shown in Section~\ref{sec:mlqa}.
\end{itemize}

\section{USER STUDY}
\label{sec:user}

\subsection{Data Sources}
For the user study we employ the two following recent MRC Spanish translations: 

\textbf{TAR:} prepared following the Translate-Align-Retrieve methodology which implies a lot of post-processing to improve the translation quality~\cite{DBLP:journals/corr/abs-1912-05200}. TAR SQuAD is produced from original English SQuAD corpus and contains both 1.1 and 2.0 versions. Further, each version contains datasets of two sizes, i.e., \emph{regular} (or default) and \emph{small} (half the size of the regular) that is less noisy and more refined.

\textbf{MT:} Stanford Question Answering Dataset (SQuAD) versions 1.1 and 2.0 translated by a private European NMT company.~\footnote{Link: \url{https://github.com/migalkin/SQuAD-es-mt}} 

SQuAD 1.1 is a reading comprehension dataset consisting of more than 100k questions posed by crowdworkers on a set of Wikipedia articles, where the answer to each question is a segment of text (span) from the corresponding passage. SQuAD 2.0 combines existing SQuAD 1.1 data with additional more than 50k unanswerable questions written by crowdworkers to look similar to the answerable ones. To do well on SQuAD 2.0, systems must not only answer questions when possible but also determine when there is no answer for the paragraph and refrain from answering.

\subsection{Translation Evaluation}
To estimate the quality of translation, 50 parallel 
examples from translated SQuAD 1.1 dataset were selected randomly. Twelve Spanish speaking evaluators were asked to give the following grades to 25 parallel examples each: 

\begin{itemize}
    \item 2, if understandable and there are only minor mistakes;
    \item 1, if understandable and has a  few major mistakes;
    \item 0, if not understandable and has more than a few major mistakes.
\end{itemize}  

In Table~\ref{table:eval} the average translation evaluation score is depicted, from which we conclude that TAR translation is significantly better than MT translation. 

\begin{table}
\centering
\caption{Translation evaluation average.}
\label{table:eval}
\begin{tabular}{l|l}
\multicolumn{1}{c|}{Translation by} & \multicolumn{1}{c}{Average}\\ \midrule
TAR & 1.717 \\
MT & 1.320 \\
\end{tabular}
\end{table}

\begin{figure}[h]
    \centering
    \includegraphics[width=\columnwidth]{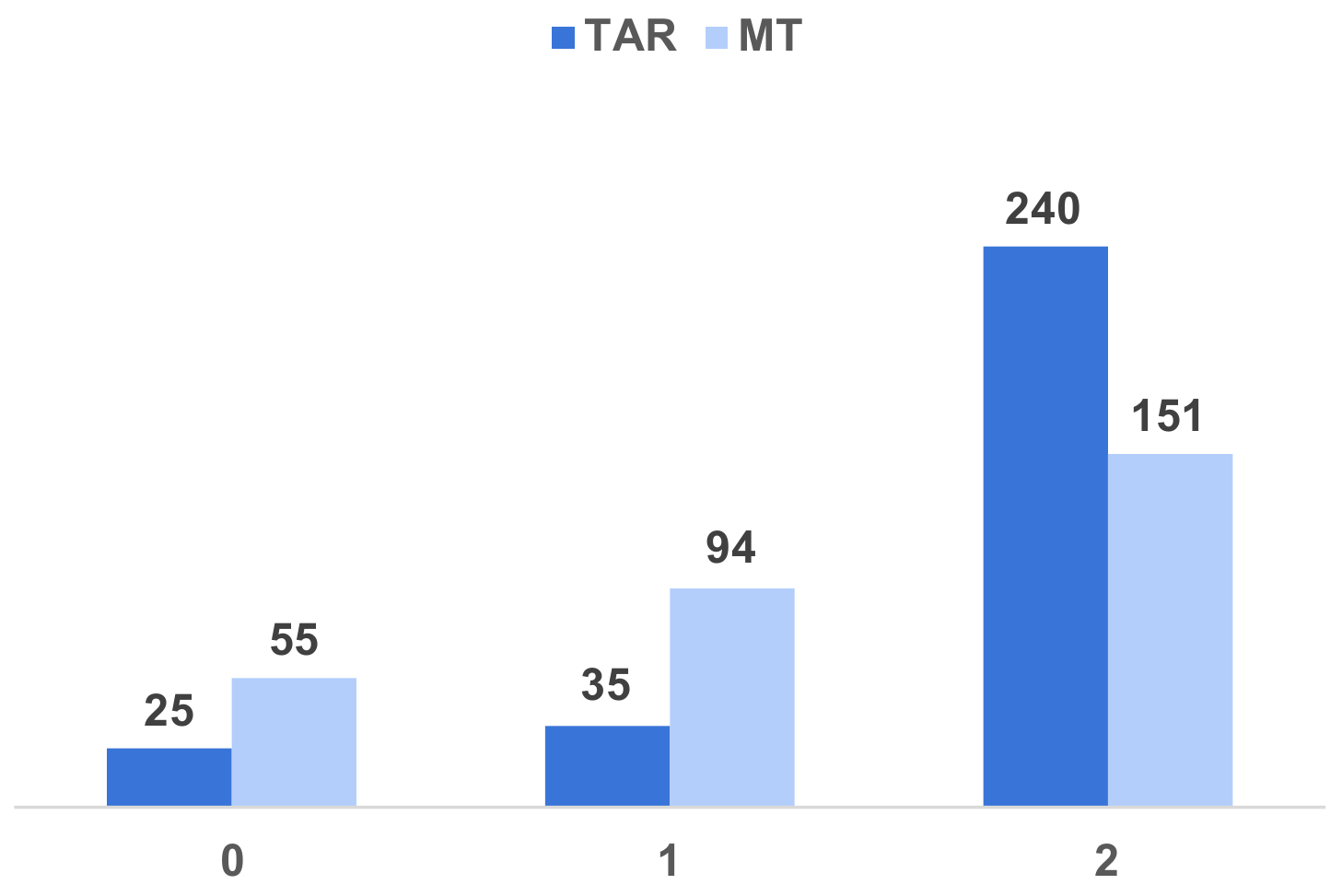}
    \caption{Score frequencies for TAR and MT translations.}
    \label{fig:hist}
\end{figure}

To inspect further, in Figure~\ref{fig:hist} we provide the histogram of score frequencies where TAR translation produces 80\% of the best scores while MT's produces only around 50\%.

\begin{table*}[t]
\centering
\caption{Performance on es-SQuAD. All models are in the case-sensitive mode. Best column results in \textbf{bold}, second best \underline{underlined}.}
\label{table:squad}
\adjustbox{max width=\linewidth}{
\begin{tabular}{@{}l|llllllll|llll@{}}
\toprule
\multicolumn{1}{c}{\multirow{4}{*}{Model}} & \multicolumn{8}{c|}{es-SQuAD (TAR)} & \multicolumn{4}{c}{es-SQuAD (MT)} \\ \cmidrule(l){2-13} 
\multicolumn{1}{c}{} & \multicolumn{4}{c|}{1.1} & \multicolumn{4}{c|}{2.0} & \multicolumn{2}{c|}{1.1} & \multicolumn{2}{c}{2.0} \\ \cmidrule(l){2-13} 
\multicolumn{1}{c}{} & \multicolumn{2}{c|}{Small} & \multicolumn{2}{c|}{Default} & \multicolumn{2}{c|}{Small} & \multicolumn{2}{c|}{Default} & \multicolumn{2}{c|}{Default} &  \multicolumn{2}{c}{Default} \\
\multicolumn{1}{c}{} & \multicolumn{1}{c}{EM} & \multicolumn{1}{c|}{F1} & \multicolumn{1}{c}{EM} & \multicolumn{1}{c|}{F1} & \multicolumn{1}{c}{EM} & \multicolumn{1}{c|}{F1} & \multicolumn{1}{c}{EM} & \multicolumn{1}{c|}{F1} & \multicolumn{1}{c}{EM} & \multicolumn{1}{c|}{F1} & \multicolumn{1}{c}{EM} & \multicolumn{1}{c}{F1} \\ 
\midrule[\heavyrulewidth]
mBERT (1.1 Sm TAR) & \textbf{57.45} & 73.34 & 55.04 & 72.38 & - & - & - & - & 56.24 & 71.12 & - & -  \\
mBERT (1.1 Def TAR) & 56.30 & 73.71 & \underline{59.65} & \underline{76.32} & - & - & - & - & 54.80 & 71.67 & -  & - \\
mBERT (2.0 Sm TAR) & \underline{57.36} & \underline{73.52} & 55.20 & 72.56 & \textbf{59.85} & 66.30 & \underline{60.18} & 66.94 & 55.36 & 70.64 & 46.17 & 57.87 \\
mBERT (2.0 Def TAR) & 56.24 & 73.11 & 59.05 & 75.48 & \underline{59.76} & \textbf{67.17} & \textbf{62.08} & \textbf{68.90} & 54.47 & 71.06 & 46.97 &  59.47 \\
mBERT (1.1 MT) & 54.25 & 71.42 & 53.90 & 71.90 & - & - & - & - & \underline{61.20} & 64.19 & - & - \\
mBERT (2.0 MT) & 52.92 & 70.51 & 53.03 & 71.25 & 29.02 & 38.60 & 26.32 & 35.80 & \textbf{61.27} & \textbf{74.15} & \textbf{61.46} & \textbf{74.72} \\
BETO & 56.72 & \textbf{74.38} & \textbf{59.71} & \textbf{76.92} & 58.55 & \underline{67.16} & 60.01 & \underline{67.97} & 55.93 & \underline{72.56} & \underline{49.49} & 62.88 \\
DistilledBETO & 54.11 & 72.41 & 57.32 & 74.94 & 57.26 & 66.28 & 58.58 & 66.75 & 52.11 & 69.86 & 48.28 & \underline{63.58} \\
\bottomrule
\end{tabular}}
\end{table*}

Furthermore, to evaluate the agreement among raters, we aggregate the evaluations in heat map representation in Figure~\ref{fig:tar_w} and Figure~\ref{fig:mt_h}. We can observe that most of the evaluators not only favoured but also synchronized well on the TAR scores, whereas MT scores appear to be more contrasting. This could possibly mean that the MT errors were so diverse that evaluators found the provided scale to some degree misleading. Hence, it was difficult to strictly represent the difference between minor and major errors.

Translation errors which could significantly affect the results have been collected and some examples are depicted in Table~\ref{table:errors}. The error types are the following: wrong gender inference, inaccurate translation or capitalization in named entities, adjectives misplacement regarding the noun. Here, we would like to point at the following errors in MT translation:

\begin{itemize}
    \item an example of combining named entity's \emph{"Warner Brothers"} in a literal Spanish translation as well as in the original state: \emph{"... y hermanos Warner. Universal, Warner Brothers ..."};
    \item an example of translation \emph{"Universal Pictures"} by changing into plural form and dropping the noun \emph{"universales"};
    \item an example of \emph{"the US War Department"} translation as an \emph{"al Departamento de Guerra DE NOSOTROS"}, an impressive translation of a capitalized abbreviation of the United States.
\end{itemize}

Therefore, we can positively answer \textbf{RQ 1} as there indeed exists a substantial difference in corpora quality when applying additional post-processing over direct machine translated data.

\section{EXPERIMENTAL STUDY}
\label{sec:exp}
In the experimental study we evaluate the performance of Spanish LMs in machine reading comprehension tasks fine-tuning them on language corpora obtained via machine translation and translation with further rule-based post-processing. \\

\textbf{Datasets.} 
For fine-tuning the pre-trained LMs in Spanish we leverage different versions of es-SQuAD, i.e., translated TAR and MT datasets described in Section~\ref{sec:user}. Small and Default versions of es-SQuAD (TAR) are annotated as \emph{sm} and \emph{def}, respectively.
For benchmarking, we employ dev sets of es-SQuAD datasets as well as test sets of MultiLingual Question Answering (MLQA)~\cite{DBLP:journals/corr/abs-1910-07475} and Cross-lingual Question Answering Dataset (XQuAD)~\cite{DBLP:journals/corr/abs-1910-11856} where both context and question are in Spanish. 

MLQA is a multi-way aligned extractive QA evaluation benchmark containing QA instances in seven languages: over 12K instances in English and 5K in each other language, with each instance parallel between 4 languages on average.
XQuAD is a benchmark dataset for evaluating cross-lingual question answering performance. The dataset consists of a subset of 240 paragraphs and 1190 question-answer pairs from the development set of SQuAD 1.1 together with their professional parallel translations into ten languages.

The systems under test are not supposed to get trained on XQuAD and MLQA. Instead, the authors of those datasets suggest to only use them as an evaluation benchmark for multi-lingual transfer learning approaches.
\\

\textbf{Models.}
We choose the pre-trained \emph{mBERT-base-cased} for fine-tuning on es-SQuAD datasets. 
For a broader comparison we also employ already pre-trained and fine-tuned on SQuAD 2.0 Spanish-only LMs BETO~\cite{CaneteCFP2020} and its distilled version DistilledBETO. 
All models allow for a maximum input sequence length of 512 tokens. Contexts of the selected datasets that do not fit into this limit are pruned to 512 tokens.
\\

\textbf{Fine-Tuning Setup.}
The models are trained and evaluated in the cased mode using the HuggingFace Transformers~\cite{Wolf2019HuggingFacesTS} framework. When fine-tuning the default hyperparameters are used: three epochs of the Adam optimizer with an initial learning rate of 0.00005.
The experiments are conducted on the Ubuntu 16.04 server equipped with one GTX 1080 Ti GPU and 256 GB RAM. 
\\

\textbf{Metrics}.
We measure EM and F1 scores in each experiment as reported by task-specific evaluation scripts. 
EM stands for exact matching of the predicted sequence with the gold label, whereas F1 measures soft matching as a harmonic mean of precision and recall of a predicted span compared to the gold span.
For consistency reasons, we do not evaluate models fine-tuned on SQuAD 1.1 datasets against SQuAD 2.0 versions.

\subsection{SQuAD Performance}
\label{sec:squad}

In the first experiment, we fine-tune mBERT on TAR and MT datasets and evaluate their accuracy on the dev test of the respective tasks. 
That is, in order to study the impact of the fine-tuning dataset we optimize the model on TAR, but evaluate on the MT dev set, and vice versa. 
The empirical results are shown in Table~\ref{table:squad}.

First, we observe that LMs fine-tuned on the MT SQuAD considerably outperform other models in terms of both EM and F1 only on the MT dev set while being significantly inferior to all other models on the TAR dev test.
For instance, mBERT fine-tuned on the MT-version of SQuAD 2.0 is about 12 EM and F1 points better than BETO on the MT-version of SQuAD 2.0 and at the same time is about 32 EM and F1 points worse than BETO in the default TAR-version of SQuAD 2.0.

Similarly, mBERT trained on the MT-version of SQuAD 1.1 achieves very good EM score on the MT-version dev set of SQuAD 1.1 but performs poorly on the TAR versions. 
Considering the difference in datasets quality demonstrated in Section~\ref{sec:user}, we deem that such a behavior is a sign of LMs sensitivity to artificially created corpora with numerous syntactic and semantic mistakes.

Moreover, the TAR-trained models show more consistent scores across the given tasks thus supporting the \textbf{RQ 2}, i.e., LMs tend to be more robust when trained and evaluated on well-prepared language corpora. Overall, in this experiment we find that LMs trained on Spanish-only corpora (e.g., BETO) perform on par or slightly better than massive multilingual LMs like mBERT fine-tuned on a similar task in a language-specific setting.

\subsection{MLQA and XQuAD Performance}
\label{sec:mlqa}

\begin{table}
\centering
\caption{Performance on MLQA and XQuAD. Cased models. Best in \textbf{bold}, second best \underline{underlined}.}
\label{table:mlqa}
\adjustbox{max width=\linewidth}{
\begin{tabular}{@{}l|llll@{}}
\multicolumn{1}{c}{\multirow{2}{*}{Model}} & \multicolumn{2}{c}{MLQA} & \multicolumn{2}{c}{XQuAD} \\ \cmidrule(l){2-5} 
\multicolumn{1}{c}{} & \multicolumn{1}{c}{EM} & \multicolumn{1}{c}{F1} & \multicolumn{1}{c}{EM} & \multicolumn{1}{c}{F1} \\ \midrule
mBERT (1.1 Sm TAR) & 42.74 & 64.36 & 53.61 & 72.89 \\
mBERT (1.1 Def TAR) & 43.14 & \underline{66.44} & 54.62 & 75.30 \\
mBERT (2.0 Sm TAR) & 43.31 & 65.04 & 54.45 & 74.09 \\
mBERT (2.0 Def TAR) & 43.44 & 66.09 & 55.97 & \underline{76.82} \\
mBERT (1.1 MT) & \underline{44.43} & 64.83 & \textbf{57.14} & 75.46 \\
mBERT (2.0 MT) & 44.13 & 64.43 & 54.03 & 73.17 \\
BETO & \textbf{45.12} & \textbf{68.77} & \underline{56.97} & \textbf{78.15} \\
DistilledBETO & 42.41 & 66.06 & 55.46 & 75.84 \\
\end{tabular}}
\end{table}

In the second experiment, we probe the TAR and MT fine-tuned models against MLQA and XQuAD in the Spanish context - Spanish question settings. 
The results are presented in Table~\ref{table:mlqa}.
A clear winner is BETO which is pre-trained on Spanish-only corpora and outperforms nearest contenders by about 2 F1 points. 
We then observe that TAR-trained models perform consistently better  than MT-trained models in terms of F1 scores.
Interestingly, in terms of EM scores mBERT 1.1 MT yields better performance than TAR and even language-specific models like BETO.
Such a phenomena can be explained by robustness of large-scale multilingual LMs that might tend to generalize better over translation artifacts. We leave further research of this phenomena to the future work.

Overall, discussing the \textbf{RQ 3} we hypothesize that for downstream language-specific tasks LMs pre-trained in that specific language are more preferable.
In case such a large-scale pre-training corpora is not available, well-processed machine translated sources tend to produce more robust LMs compared to purely machine translated sources.

\section{CONCLUSION AND FUTURE WORK}
In this work, we studied the impact of machine translated corpora quality on question answering tasks. 
Having formulated three research questions, we employed Spanish SQuAD-style datasets for empirical evaluation.
User study confirmed there is a significant difference in dataset quality and amount of language artifacts. 
Further experimental studies confirmed that LMs are sensitive to the quality of machine translated corpora. 
We also observe signs of LMs robustness to translation defects in downstream transfer learning tasks.

For the future work we pose a question towards conducting an appropriate analysis on how neural networks overcome the flaws in the data, being not always machine translated, to become robust and noise resilient. 

\bibliography{ecai}

\clearpage

\begin{table*}[]
\centering
\caption{A sample of TAR and MT translations error examples annotated in the user study.}\label{tab1}
\label{table:errors}

\begin{tabularx}{\linewidth}{Y Y Y}
    English & TAR & MT\\ \hline
\vspace{1pt}    
\end{tabularx}

\begin{tabular}{l}
\multicolumn{1}{c}{\textbf{Gender inference} (TAR avg score = 0.8, MT avg score = 0) } \\
\end{tabular}
\hrule
\begin{tabularx}{\linewidth}{ X  X  X}
    It's not clear, however that this stereotypical view reflects the reality of East Asian classrooms or that the educational goals in these countries are commensurable with \hl{those in Western countries}. In Japan, for example, although average attainment on standardized tests may exceed \hl{those in Western countries}, classroom discipline and behavior is highly problematic. Although, officially, schools have extremely rigid codes of behavior, in practice many teachers find the students unmanageable and do not enforce discipline at all. & 
    
    Sin embargo, no está claro que esta opinión estereotipada refleje la realidad de las aulas de Asia oriental o que los objetivos educativos de esos países sean acordes con \hl{los de los países occidentales}. En Japón, por ejemplo, aunque el rendimiento medio de los ensayos estandarizados puede superar \hl{los de los países occidentales}, la disciplina y el comportamiento de las aulas son muy problemáticos. Aunque oficialmente las escuelas tienen códigos de conducta extremadamente rígidos, en la práctica muchos maestros consideran que los estudiantes son inmanejables y no aplican la disciplina en absoluto. & 
    
    No está claro, sin embargo, que esta visión estereotipada refleja la realidad de las aulas del Asia oriental o que los objetivos educativos en estos países son conmensurables con \hl{los de los países occidentales}. En Japón, por ejemplo, aunque el logro promedio de las pruebas estandarizadas puede exceder \hl{las de los países occidentales}, la disciplina y el comportamiento en el aula son altamente problemáticos. Aunque, oficialmente, las escuelas tienen códigos de comportamiento extremadamente rígidos, en la práctica muchos profesores encuentran a los estudiantes inmanejables y no aplican la disciplina en absoluto.\\
\vspace{1pt}
\end{tabularx}
\hrule
\begin{tabular}{l}
\multicolumn{1}{c}{\textbf{Translation and capitalization inconsistency in named entities} (TAR avg score = 1, MT avg score = 0.1)} \\
\end{tabular}
\hrule
\begin{tabularx}{\linewidth}{ X  X  X}
    The motion picture, television, and music industry is centered on the \hl{Los Angeles} in southern California. Hollywood, a district within Los Angeles, is also a name associated with the motion picture industry. Headquartered in southern California are \hl{The Walt Disney Company} (which also owns ABC), \hl{Sony Pictures}, \hl{Universal}, MGM, \hl{Paramount Pictures}, \hl{20th Century Fox}, and \hl{Warner Brothers}. Universal, \hl{Warner Brothers}, and Sony also run major record companies as well. & 
    
    La industria del cine, la televisión y la música se centra en \hl{Los Ángeles} en el sur de California. Hollywood, un distrito dentro de Los Ángeles, es también un nombre asociado a la industria cinematográfica. Con sede en el sur de California están \hl{The Walt Disney Company} (que también posee ABC), \hl{Sony Pictures}, \hl{Universal}, MGM, \hl{Paramount Pictures}, \hl{20th Century Fox}, y \hl{Warner Brothers}. Universal, \hl{Warner Brothers} y Sony también tienen grandes compañías discográficas. & 
    
    La imagen del movimiento, la televisión y la industria musical se centran en \hl{los Ángeles} en el sur de California. Hollywood, un distrito de los Ángeles, es también un nombre asociado a la industria fotográfica de movimiento. Con sede en el sur de California están \hl{la compañía Walt Disney} (que también posee ABC), \hl{imágenes de Sony}, \hl{universales}, MGM, \hl{imágenes principales}, \hl{Fox del siglo 20} y \hl{hermanos Warner}. Universal, \hl{Warner Brothers} y Sony también dirigen grandes empresas de registro.\\
\vspace{1pt}
\end{tabularx}

\begin{tabularx}{\linewidth}{ X X X}
    During the same year, Tesla wrote a treatise, The Art of Projecting Concentrated Non-dispersive Energy through the Natural Media, concerning charged particle beam weapons. Tesla published the document in an attempt to expound on the technical description of a "superweapon that would put an end to all war." $<$...$>$ Tesla tried to interest \hl{the US War Department}, the United Kingdom, the Soviet Union, and Yugoslavia in the device. & 
    
    Durante el mismo año, escribió un tratado, The Art of Projecting Concentrated non-dispersive Energy through the Natural Media, sobre las armas de haz de partículas cargadas. Tesla publicó el documento en un intento de exponer la descripción técnica de una "superarma que pondría fin a toda guerra". $<$...$>$ Tesla trató de interesar \hl{al Departamento de Guerra de los Estados Unidos}, el Reino Unido, la Unión Soviética y Yugoslavia en el dispositivo. & 
    
    Durante el mismo año, Tesla escribió un treatise, el arte de proyectar energía concentrada no dispersa a través de los medios naturales, en relación con las armas de haz de partículas cargadas. Tesla publicó el documento en un intento de exponer la descripción técnica de un «superarma que pondría fin a toda guerra». $<$...$>$ Tesla trató de interesar \hl{al Departamento de Guerra DE NOSOTROS}, al Reino Unido, a la Unión Soviética y a Yugoslavia en el dispositivo.\\
\vspace{1pt}
\end{tabularx}
\hrule
\begin{tabular}{l}
\multicolumn{1}{c}{\textbf{Adjectives placement regarding the noun} (TAR avg score = 1, MT avg score = 0) } \\
\end{tabular}
\hrule
\begin{tabularx}{\linewidth}{ X X X}
    CBS broadcast Super Bowl 50 in the U.S., and charged an average of \$5 million for \hl{a 30-second commercial} during the game. The Super Bowl 50 halftime show was headlined by the British rock group Coldplay with special guest performers Beyoncé and Bruno Mars, who headlined the Super Bowl XLVII and Super Bowl XLVIII halftime shows, respectively. It was the third-most watched U.S. broadcast ever. &
    
    CBS transmitió el Super Bowl 50 en los Estados Unidos, y cobró un promedio de \$5 millones por \hl{un comercial de 30 segundos} durante el juego. El espectáculo de medio tiempo del Super Bowl 50 fue encabezado por el grupo de rock británico Coldplay con artistas invitados especiales como Beyoncé y Bruno Mars, quienes encabezaron los shows de medio tiempo del Super Bowl XLVII y Super Bowl XLVIII, respectivamente. Fue el tercer programa más visto de Estados Unidos. & 
    
    CBS emitió 50 super bowl en los U. S. y cobró un promedio de US \$5 millones por \hl{un 30 - segundo comercial} durante el juego. El espectáculo de semáforo 50 de super fue encabezado por el grupo de rock británico Coldplay con artistas invitados especiales Beyoncé y Bruno Mars, que encabezaron el súper súper XLVII y los espectáculos de semestral XLVIII, respectivamente. Fue la tercera, la más observada u.\\
\vspace{1pt}
\end{tabularx}

\end{table*}

\end{document}